# Can We Enable the Drone to be a Filmmaker?

Yuanjie Dang

*Abstract*—Drones are enabling new forms of cinematography. However, quadrotor cinematography requires accurate comprehension of the scene, technical skill of flying, artistic skill of composition and simultaneous realization of all the requirements in real time. These requirements could pose real challenge to drone amateurs because unsuitable camera viewpoint and motion could result in unpleasing visual composition and affect the target's visibility. In this paper, we propose a novel autonomous drone camera system which captures action scenes using proper camera viewpoint and motion. The key novelty is that our system can dynamically generate smooth drone camera trajectory associated with human movement while obeying visual composition principles. We evaluate the performance of our cinematography system on simulation and real scenario. The experimental results demonstrate that our system can capture more expressive video footage of human action than that of the state-of-the-art drone camera system. To the best of our knowledge, this is the first cinematography system that enables people to leverage the mobility of quadrotor to autonomously capture high-quality footage of action scene based on subject's movements.

## I. INTRODUCTION

The emergence of drones has raised the bar for cinematic quality and visual storytelling for filmmakers. With a continuous decrease in the cost of drones, more and more people can get the hands on them to capture stunning visuals. While recent advancements in technology have made aerial videography more approachable, cinematography using a drone is not a very easy activity. For example, to capture a high-quality footage of human action, drone cinematography requires several professional techniques: accurate comprehension of the dynamic scenarios, practiced skill of flying, artistic skill of composition and simultaneously realization of all the requirements in real time. This technical threshold results in the gap between experts and amateurs. Many videos taken from amateurs suffer from the following problems: 1) non-smooth camera motion results in unpleasing footage. 2) unsuitable viewpoint affects the saliency and visibility of the moving subject.

Autonomous drone systems for capturing footage of moving subjects appeared in the recent years. The most popular function is "Follow-Me" which utilizes the front camera to track the moving target. The camera control is driven by the target's position in the screen space. This approach is suitable to follow a moving target in a large-scale outdoor environment. However, if we shoot a video of complex human movement (e.g. ballet dance), the active track suffers from the disadvantage that the footage captured from a drone with single viewpoint and single-axis motion cannot satisfy the human's aesthetic needs. Another filming approach is based on the sensors on the subject. This approach localizes the subject with a portable GPS remote sensor. With the subject's GPS information, the drone can customize the shooting viewpoints with multiple-axes camera motion (e.g. orbiting). However, the inaccurate GPS localization could result in inconsistent visual composition, even track loss.

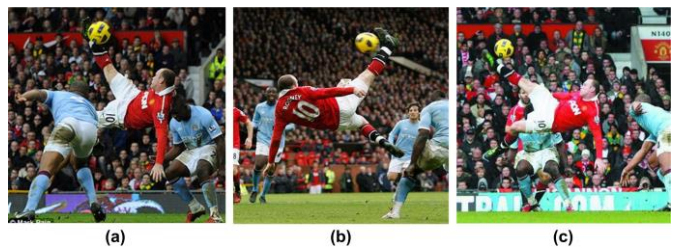

Fig.1. The motion capture from different viewpoints.

Some methods in computer graphics, such as virtual camera control in 3D animation, have been introduced to the state-of-the-art drone cinematography. In general, existing methods mainly focus on two critical issues for an automated drone camera system: 1) rules for evaluating the quality of the captured video, which can then be used to guide automatic path planning and viewpoint selection, and 2) methods for generating feasible and smooth trajectories. Joubert et al. [1] create a semi-autonomous quadrotor camera system that positions a drone relative to the people in a scene according to the Rule of Third to guide the path planning and viewpoint selection. In addition, they use the gap between the object's position in the camera screen and the golden point to evaluate the viewpoint quality. In computer graphics, there are some other metrics [2, 3, 4] to quantify the viewpoint quality such as limb's visibility. These viewpoint quality metrics can be also used to evaluate the footage of human action. For example, Fig.1 shows the motion capture from three different viewpoints. Fig.1(c) has highest viewpoint quality in terms of the visibility of the body parts. In addition, camera trajectory planning is also significant for visual perception. Sudden or jittering camera motion will result in unpleasing footage. Some techniques have been proposed to automatically generate high-quality aerial video along with smooth trajectory. For example, Roberts et al. [20, 21] develop an interactive tool "Horus" to generate dynamically feasible trajectories for capturing aerial video of static scenarios. This work is based on a lightweight library "Flashlight" [21] for analyzing and solving the control problem. Huang et al [25-30] proposed a learning-based drone cinematography system for action scenes.

In general, these systems only consider the target's placement on the image or GPS position. Few studies can provide a drone filming method based on subject's movement. Our work is to build a novel autonomous drone cinematography system to capture human action. The key

Yuanjie Dang, is with the College of Information and Engineering, Zhejiang University of Technology, Hangzhou 310023 China. (e-mail: dangyj@zjut.edu.cn).

novelty is that our system can dynamically generate smooth drone camera trajectory associated with human movement while obeying visual composition principles. To the best of our knowledge, we are first to propose an autonomous drone camera system for capturing action scenes with the aesthetic viewpoint based on the subject's movement.

## II. RELATED WORK

The autonomous drone filming techniques can date back to the "Follow-Me" [5, 6, 7, 8]. The drone utilizes its embedded front camera to track the moving target. Celine et al. [5] presented a vision-based algorithm to autonomously track and chase a moving target with a small-size flying UAV. The proposed approach to estimate the target's position, orientation and scale, is built on a robust color-based tracker using a multi-part representation. Coaguila et al. [8] considered the problem of moving a quadrotor camera to capture the frontal facial video of subjects based on subject's facing.

An alternative approach is to place sensors on subjects, relieving the quadrotor from maintaining a visual line of sight to all subjects. Given 2D position of subjects, the drone can shoot video by using multiple axes of motion (e.g. orbiting), which makes the video more dynamic. [1, 9] leverage centimeter-accurate RTK GPS to track human movement. Niels et al. [1] combine IMU (Inertial Measurement Units) and RTK GPS to track subject and demonstrate its efficacy for automating cinematography. In addition, they add the visual composition principle to guide the camera control. Although the system has been successfully used to film a range of activities, such as taking a selfie, playing catch, receiving a diploma, the subjects are fairly stationary and the camera control does not respond to the limbs movements. Therefore, the camera trajectory and viewpoint selection is computed offline, which is inconsistent with real-time motion tracking.

In cinematography and computer graphics, camera control is a well-established problem focusing on searching for a suitable camera configuration for capturing a scene narrative, while obeying a set of cinematographic rules [11], as well as other constraints such as occlusion [16], objects visibility [17], layout in the resulting image [12], and orientations [13]. Using these attributes the system measures the quality of each frame taken from different viewpoint and outputs the best view. Tomlinson et al. [14] propose a behavior-based autonomous cinematography system by encoding the camera as a creature with motivational desires, focusing on camera movement styles and lighting in order to augment emotional content. Halper et al [15] presents a real-time automatic, dynamic camera control for video games to manage trade-off between constraint satisfaction and frame coherence.

## III. METHOD

The framework of our autonomous drone cinematography system is illustrated in Fig. 2. First, we use depth camera to shoot the action scene to get the subject's depth. Second, we predict the next best viewpoint based on the subject's 3D movement, which guides the subsequent dynamic trajectory planning. Third, the drone flies to the desired waypoint and capture the image. In addition, we incorporate a camera adjustment module to compensate for errors caused by trajectory drift.

This paper considers scenarios with only one subject. In the following sections, we offer a detailed discussion on each module, namely, viewpoint estimation (Sec.3.1), trajectory planning (Sec. 3.2), and camera adjustment (Sec.3.3).

### A. Viewpoint Estimation

In this section, we introduce viewpoint estimation. First, the depth camera captures the 3D information of the moving subject. In our experiment, we use Kinect sensor, which identifies human movement with 25 skeletal joints. To simplify the computation, we select 13 major points, which captures the head, the spine shoulder, the spine base, the left and right shoulders, both elbows, hands, knees and feet, to represent one pose (see Fig.3 (left)). For each frame, we assess and analyze viewpoint quality in the polar coordinate centered in subject (see Fig.3 (right)), and we search the camera pose with the highest viewpoint quality as optimal viewpoint.

We only consider the viewpoint estimation in terms of orientation, so we fix the distance between subject and the camera at 2.5m and the height at 2.2m. Assa et al [2] selects the expressive viewpoint of 3D animation based on the motion and visibility of the body parts. They measure the viewpoint quality by summing a set of quality descriptors. This operation, however, fails to achieve the ideal result as it cannot distinguish different action scenes.

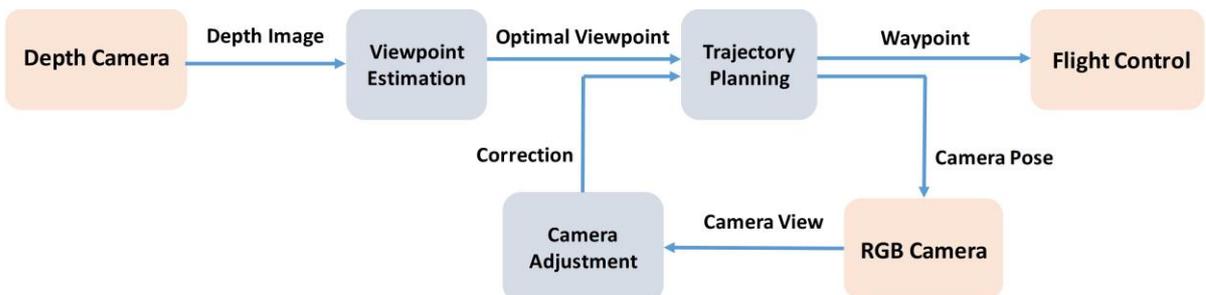

Fig. 2. Pipeline

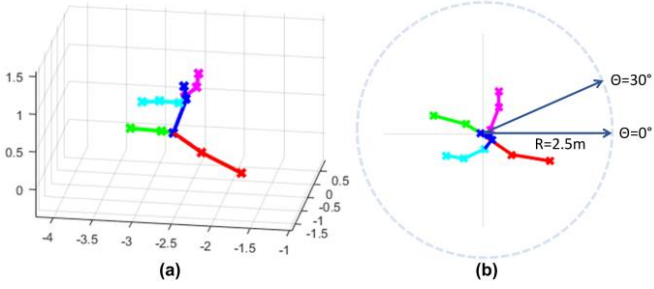

Fig.3. (left) The 3D human skeleton. (right) Top view in the subject-center polar coordinate.

As Fig.4 (a) shows, when the subject stands still or moves at a low speed, the ideal viewpoint maximizes the visibility of the subject. When the subject moves fast, Fig.4(b) provides a dynamic visual effect by placing the camera perpendicular to the movement direction of the subject.

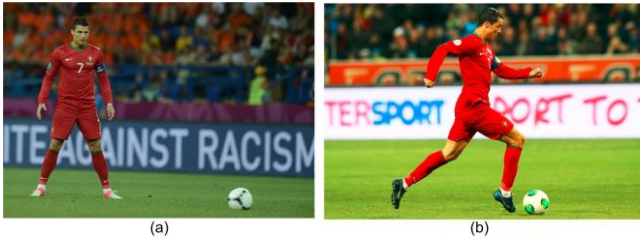

Fig.4. (a) When the subject stands still, this viewpoint provides maximum limbs' visibility to help viewer focus on the subject. (b) When the subject moves at a high speed, the viewpoint perpendicular to the movement direction can give better dynamic visual effect.

Based on the discussion above, we present an automatic viewpoint estimation based on two descriptors: velocity perpendicular descriptor and projection area descriptor. Velocity perpendicular descriptor is used to measure the difference between the current viewpoint and the perpendicular vector of the displacement of the subject's centroid in horizontal plane. Projection area descriptor describes the scatter of the skeleton joint points on the camera screen, and we use the variance of 2D location of projected skeleton joints to measure the projected area. If we use both descriptors to measure the viewpoint quality, we can see that the velocity perpendicular descriptor demonstrates a bimodal triangular distribution in the orientation of the camera while the projection area descriptor presents a multimodal distribution, as are shown in Fig. 5. Both distributions have translational symmetry with respect to 180 degrees.

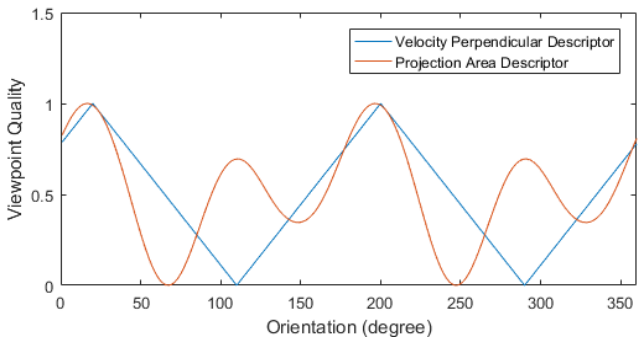

Fig.5. Normalized viewpoint quality in terms of velocity perpendicular descriptor and projection area descriptor for a given subject's movement.

The drone determines the optimal viewpoint in the following procedure: first, it calculates the subject's displacement. Second, if the displacement is smaller than the predefined threshold, the drone will position itself in the angle that enables it to shoot pictures containing the maximum projection of the subject; if the placement is larger than the predefined threshold, it will position itself in the angle that is perpendicular to the direction of the subject's movement.

### B. Trajectory Planning

Assuming that the human action between adjacent frames is subtle, we estimate the next optimal viewpoint based on the current subject's movement. The camera view captured from the optimal viewpoint has the highest image quality. However, the camera trajectory that goes through these viewpoints tends to lead to the camera jumping around too much. In addition, because we publish the waypoint (i.e. optimal viewpoint) to the flight control periodically, the drone is required to reach the goal before the new waypoint is published. However, the dynamic constraint of the drone does not allow the camera to move with too large linear or angular speed, which means that the drone is not likely to arrive at each optimal viewpoint within a command cycle $T$. This subsection we pay particular attention to a trade-off between constraint satisfaction and viewpoint quality.

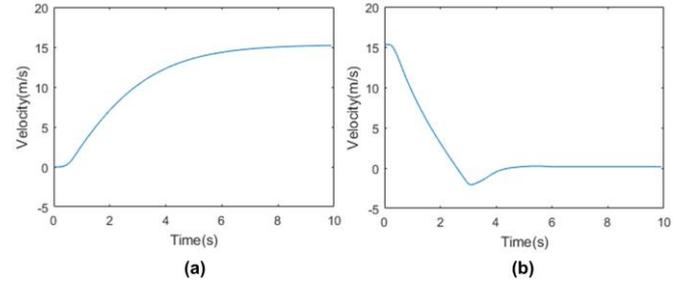

Fig. 6. The velocity-time diagram in (a) acceleration and (b) deceleration. The different roles of wind resistance results in that acceleration process is not the reverse process of deceleration. The inertia of the drone causes subtle oscilation around zero speed at the end of deceleration.

A practical method is to detect a local optimal viewpoint within a feasible action range as the waypoint. In fact, it is hard to define an accurate range for the action region for the following reasons: 1) The action region is determined by the initial velocity, maximum speed and acceleration of the drone. 2) The acceleration process is not the reverse process of deceleration (see Fig. 6). 3) The drone performs variably accelerated motion.

Based on the above discussion, we define the action region as the sum of two segments of arcs, namely, the feasible flight distance during deceleration $s_{dec}$ and that during acceleration $s_{acc}$ in a given time period (see Fig.7).

$$0 \leq s_{dec} \leq s_{acc} \leq v_{\max} T \qquad (1)$$

Eqn.1 describes the relation between $s_{dec}$ and $s_{acc}$. $s_{dec}$ and $s_{acc}$ are determined by the combination of the initial velocity, wind resistance, thrust force, the drone's weight and command cycle. In order to simplify the computation, we set both $s_{dec}$ and $s_{acc}$ as $\frac{1}{2} v_{max} T$.

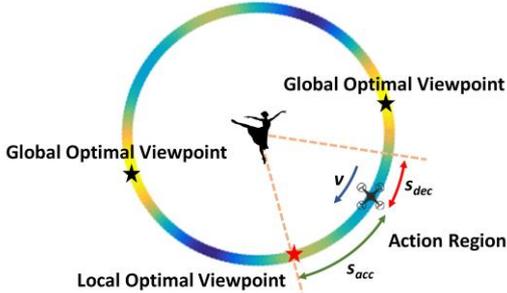

Fig.7. Given a subject's pose, we visualize the viewpoint quality in the orientations of the camera as a quality map (the yellow region represents the high viewpoint quality while the blue region represents the low quality). To estimate the next waypoint, the drone detects the local optimal viewpoint within a limited action region, which consists of the two segments of feasible flight distances during deceleration $s_{dec}$ and that during acceleration $s_{acc}$.

Given a limited action region, we perform the gradient descent method to detect the local optimal viewpoint as the next waypoint as Fig. 7 shows. Because each waypoint is dynamically generated, the overall flight trajectory is not smooth. In order to smooth the camera trajectory, we consider a recent history of the waypoints to correct the position of waypoint as

$$P_t = (1 - \alpha)P_{t-1} + \alpha D_t(n) \quad (2)$$

where $D_t$ is defined as the predicted waypoint (i.e. local optimal viewpoint). $\alpha$ is the smoothness factor.

Once the drone receives the next waypoint command, it moves to the waypoint while keeping the safe distance from the subject. The drone will position itself to face the subject. Under the ideal condition, this pose places the subject in the center of the image.

*C. Camera Adjustment*

If the drone can reach each waypoint at a given time, the camera can achieve perfect alignment with the desired trajectory and capture the footage with consistent visual composition. However, the drone fails to reach the desired waypoint accurately. There are some factors that affects the flight trajectory: 1) the accumulated errors caused by the Inertial Measurement Unit (IMU) embedded on the drone results in trajectory drift. 2) The position accuracy of drone GPS is limited (within 1.5m). 3) Windy weather also causes the trajectory drift. These factors result in the failure that we do not manage to frame a subject accurately, since small errors in orientation can significantly impact the visual composition. Therefore, the vision-based feedback is crucial to shoot pleasing video. More specifically, the target position in the image will be utilized to adjust the camera orientation.

In our system, we set the desired target position as the center of the image. Assumed that only one subject moves in a static scene, we use motion detection from moving camera [20] to detect and track the subject. The approach [20] outputs a binary mask that represents the region of the subject. We calculate the difference between the center of the mask and the center of the image as the input of PID controller. The controller attempts to correct the visual composition over time by orientation adjustment. As Fig. 8 shows, the green line is the desired trajectory and the red line is the actual trajectory. The drone is supposed to reach the waypoint *A* at a given time, but it moves to the waypoint *B* because of the trajectory drift. At this time, the visual composition error is detected. The drone adjusts the camera orientation at next waypoint *D* to minimize the gap between the target and actual position. Compared with the image captured in the desired waypoint *C*, the camera view at waypoint *D* is slight different in terms of scale, but it does not affect the user experience.

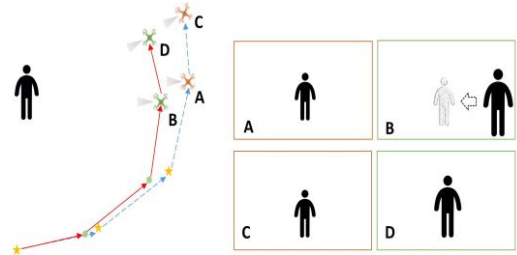

Fig. 8. Vision-based feedback can adjust the orientation to compensate the visual composition error caused by trajectory drift.

## IV. EXPERIMENTS

In this section, we will evaluate the visual effect of the footage generated from our system on dataset CMU Motion Capture Dataset and real-time Tai Chi performance. We conduct the experiments on ThinkPad T450, which runs Windows 10 with 8 GB RAM and an i7-5600U CPU at 2.6Hz. We develop the flight control system based on DJI Mavic Pro.

We compare the footage generated by our system with Follow-Me, in which the camera can autonomously keep distance to follow the target and adjust the camera to place subject on the center of the camera screen. It is noted that there are several state-of-the-art drone cinematography systems similar to us. Joubert et al [1] use a drone to automatically capture well-composed footage for action scene. However, it only works for two subjects and the subjects must stay within a fixed safety sphere during filming. Coaguila et al. [8] develop a quadrotor camera to automatically capture the frontal facial video of subjects based on subject's facing. However, our drone camera is not close enough to the subject for recovering the subject's face orientation. Roberts et al. [20, 21] develop an interactive tool "Horus" to generate smooth trajectories for capturing aerial video, but users are required to manually input all the waypoints. Based on the discussion above, we do not consider these methods as comparison.

*A. CMU Motion Capture Dataset*

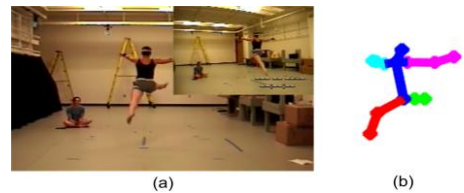

Fig. 9. CMU Motion Capture Dataset. (a) video (b) 3D skeleton.

CMU Motion Capture Dataset includes a set of the motion data and reference video (see Fig. 9). For each motion data, there is only one subject in the action scene. The 3D motion data is extracted from 41 Vicon markers taped on the subject's body. The motion data is recorded for 6-12 seconds in 120Hz. In our application, we only consider 13 markers to represent the human action for simplification. We use motion data with topic "Dance" for our testing. 11 subjects are involved and their movements include sideways steps, pirouette, glissade devant and so on.

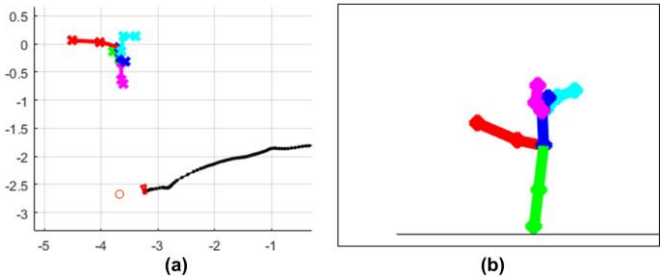

Fig. 10. (a) The topview includes the subject, camera trajectory (black line), camera orientation and optimal viewpoint (red circle). (b) The corresponding camera view.

We simulate the generated footage based on the camera orientation and position as Fig. 10 shows. We evaluate the quality of the footage in terms of two aspects: 1) visual composition. 2) viewpoint quality.

*1) Visual Composition*

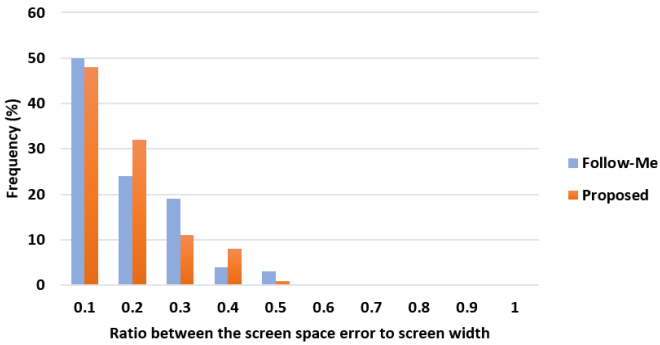

Fig. 11. The distribution of visual composition

We define the difference between actual subject's screen position and the center of the image as screen space error. We use the ratio between the screen space error and the screen width to measure visual effect of composition. Fig. 11 shows that our proposed method achieves the similar performance to Follow-Me.

*2) Viewpoint Quality*

This section we evaluate the difference in viewpoint quality between Follow-Me and our system. Viewpoint quality can be defined as the angle difference between the actual viewpoint and its global optimal viewpoint for each frame. Fig. 12 shows that our drone system can capture the footage from good viewpoint with higher frequency. This can be explained as the Follow-Me just follows the subject and ignores the pose.

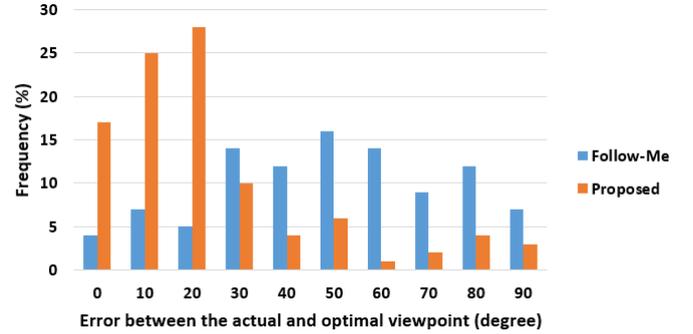

Fig. 12. The distribution of viewpoint quality

Tab.1 gives the comparison between Follow-Me and our proposed method in terms of the average screen space error ratio and the average viewpoint error. Comparing with Follow-Me, we can see that our method can achieve better viewpoint quality with a similar visual composition performance.

TABLE I

PERFORMANCE COMPARISON OF BOTH METHODS

| Method | Average Screen Space Error Ratio | Average Viewpoint Error |
|---|---|---|
| Proposed | 0.182 | 23.6° |
| Follow-Me | 0.186 | 49.3° |

*B. Real-Time Tai Chi Performance*

We choose the Tai Chi as the real test action because of its complex motion and slow transition. We run our system on a real drone DJI Mavic Pro to capture the footage (see Fig. 13). The 3D motion data is captured by a Kinect sensor fixed on the ground. We compare our proposed method with Follow-Me, which is Active Track mode of DJI drone. In our experiment, the subject starts with the same relative position to the drone and repeats the Tai Chi for both methods. Fig. 14 shows that our system can capture more pleasing footage of the action scene than Follow-Me. This improvement is expressed by better camera viewpoint and dynamic visual effect generated by the adaptive camera motion.

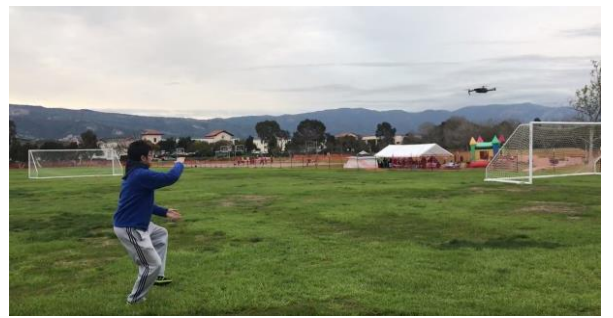

Fig. 13. The experimental setup

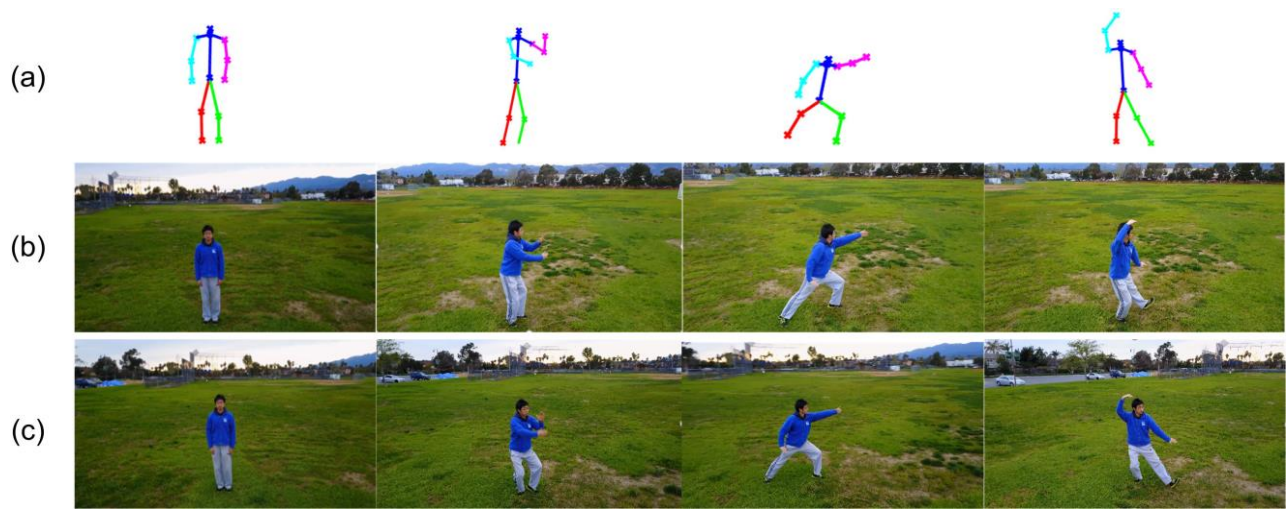

Fig. 14 (a) 3D motion data generated by Kinect. (b) The video captured by Active Track mode of DJI Go. (c) Proposed Method.

## V. CONCLUSION

We present a system that attempts to follow a set of cinematography rules when autonomously capturing footage of human action with a drone. Our novelty is that our system can dynamically generate smooth flight trajectory associated with subject's movement while shooting visually pleasing footage. We test our system on the simulation and real scenarios. The quantitative experiments on the simulation demonstrate that our system can manage the tradeoff between viewpoint quality and trajectory feasibility. We use a real drone to capture several footages in action scene, which qualitatively proves that our system achieves better visual effect than the fixed camera.

Next step is to extend our system to the large-scale action scene involving multiple subjects based on more sophisticated viewpoint quality descriptors.